\documentclass{article}

\usepackage{microtype}
\usepackage{graphicx}
\usepackage{subfigure}
\usepackage{multicol}
\usepackage{booktabs} 

\usepackage{xr-hyper}
\usepackage[bookmarksopen,
            breaklinks=true,
            colorlinks=true,
            citecolor=black,
            urlcolor=blue]{hyperref}

\usepackage[accepted]{icml2020}

\usepackage{times}
\usepackage{amssymb}        
\usepackage{enumitem}

\graphicspath{{figs/}{../figs/}}  

\usepackage{amsmath,amsfonts,bm}

\def\eqref#1{equation~\ref{#1}}

\def\1{\bm{1}}

\def\eps{{\epsilon}}

\DeclareMathAlphabet{\mathsfit}{\encodingdefault}{\sfdefault}{m}{sl}
\SetMathAlphabet{\mathsfit}{bold}{\encodingdefault}{\sfdefault}{bx}{n}

\newcommand{\paren}[1]{ \left(#1\right) }

\def\URLESAPROBAV{https://www.esa.int/Our_Activities/Observing_the_Earth/Proba-V}

\def\makeabstract{
\begin{abstract}
Generative deep learning has sparked a new wave of Super-Resolution (SR) algorithms that
enhance single images with impressive aesthetic results, albeit with imaginary details.
Multi-frame Super-Resolution (MFSR) offers a more grounded approach to the ill-posed problem, by conditioning on multiple low-resolution views.
This is important for satellite monitoring of human impact on the planet -- from deforestation, to human rights violations -- that depend on reliable imagery.
To this end, we present HighRes-net, the first deep learning approach to MFSR that learns its sub-tasks in an end-to-end fashion: (i) \textbf{co-registration}, (ii) \textbf{fusion}, (iii) \textbf{up-sampling}, and (iv) \textbf{registration-at-the-loss}.
Co-registration of low-resolution views is learned implicitly through a reference-frame channel, with no explicit registration mechanism.
We learn a global fusion operator
that is applied recursively on an arbitrary number of low-resolution pairs.
We introduce a \textit{registered loss},
by learning to align the SR output to a ground-truth through ShiftNet.
We show that by learning deep representations of multiple views, we can super-resolve low-resolution signals and enhance
Earth Observation data at scale.
Our approach recently topped the European Space Agency's MFSR competition on real-world satellite imagery.
\end{abstract}
}

\def\makeacknowledgments{
\section*{Acknowledgments}
    We would like to thank Catherine Lefebvre, Laure Delisle, Alex Kuefler, Buffy Price, David Duvenaud, Anna JungBluth, Carl Shneider, Xavier Gitiaux, Shane Maloney for their helpful comments on our manuscript.
    We are extremely grateful to the Advanced Concepts Team of the ESA for organizing the Kelvin competition, and the participating teams for working hard to pushing multi-frame super-resolution to its limits.
    Finally, this version has been improved considerably thanks to the feedback that we received during the \href{https://openreview.net/forum?id=HJxJ2h4tPr}{peer-review process of ICLR 2020}.
}

\makeatletter
\newcommand*{\addFileDependency}[1]{
  \typeout{(#1)}
  \@addtofilelist{#1}
  \IfFileExists{#1}{}{\typeout{No file #1.}}
}
\makeatother
 
\newcommand*{\myexternaldocument}[1]{%
    \externaldocument{#1}%
    \addFileDependency{#1.tex}%
    \addFileDependency{#1.aux}%
}   

\myexternaldocument{supp}

\icmltitlerunning{HighRes-net: Recursive Fusion for Multi-Frame Super-Resolution of Satellite Imagery}

\begin{document}

\twocolumn[
\icmltitle{HighRes-net: Recursive Fusion for Multi-Frame Super-Resolution \\ of Satellite Imagery}



\icmlsetsymbol{equal}{*}

\begin{icmlauthorlist}
\icmlauthor{Michel Deudon}{equal,eai}
\icmlauthor{Alfredo Kalaitzis}{equal,eai}
\icmlauthor{Israel Goytom}{mila}
\icmlauthor{Md Rifat Arefin}{mila}
\icmlauthor{Zhichao Lin}{eai}
\icmlauthor{Kris Sankaran}{mila,uom}
\icmlauthor{Vincent Michalski}{mila,uom}
\icmlauthor{Samira E. Kahou}{mila,mcg}
\icmlauthor{Julien Cornebise}{eai}
\icmlauthor{Yoshua Bengio}{mila,uom}
\end{icmlauthorlist}

\icmlaffiliation{eai}{Element AI, London, UK}
\icmlaffiliation{mila}{Mila, Montreal, Canada}
\icmlaffiliation{uom}{Universit\'{e} de Montr\'{e}al, Montreal, Canada}
\icmlaffiliation{mcg}{McGill University, Montreal, Canada}

\icmlcorrespondingauthor{Alfredo Kalaitzis}{freddie@element.ai}

\icmlkeywords{multi-frame super-resolution, super-resolution, super resolution, multi-frame, multi-image, multi image, multi frame, multi-temporal, fusion, upscaling, upsampling, up-scaling, up-sampling, remote sensing, remote-sensing, remote-sensing, dealiasing, de-aliasing, dealias, de-alias,  deep learning, registration, earth observation, satellite, satellite data, satellite imagery, computer vision, ICML, Lanczos filter, Lanczos, image registration, registration, homographynet, homography}

\vskip 0.3in
]



\printAffiliationsAndNotice{\icmlEqualContribution} 

\makeabstract


\section{Introduction}
\label{sec:1.intro}

\begin{figure*}[ht]
    \begin{center}
        \includegraphics[width=0.49\linewidth, trim={0 6ex 0 0}, clip]{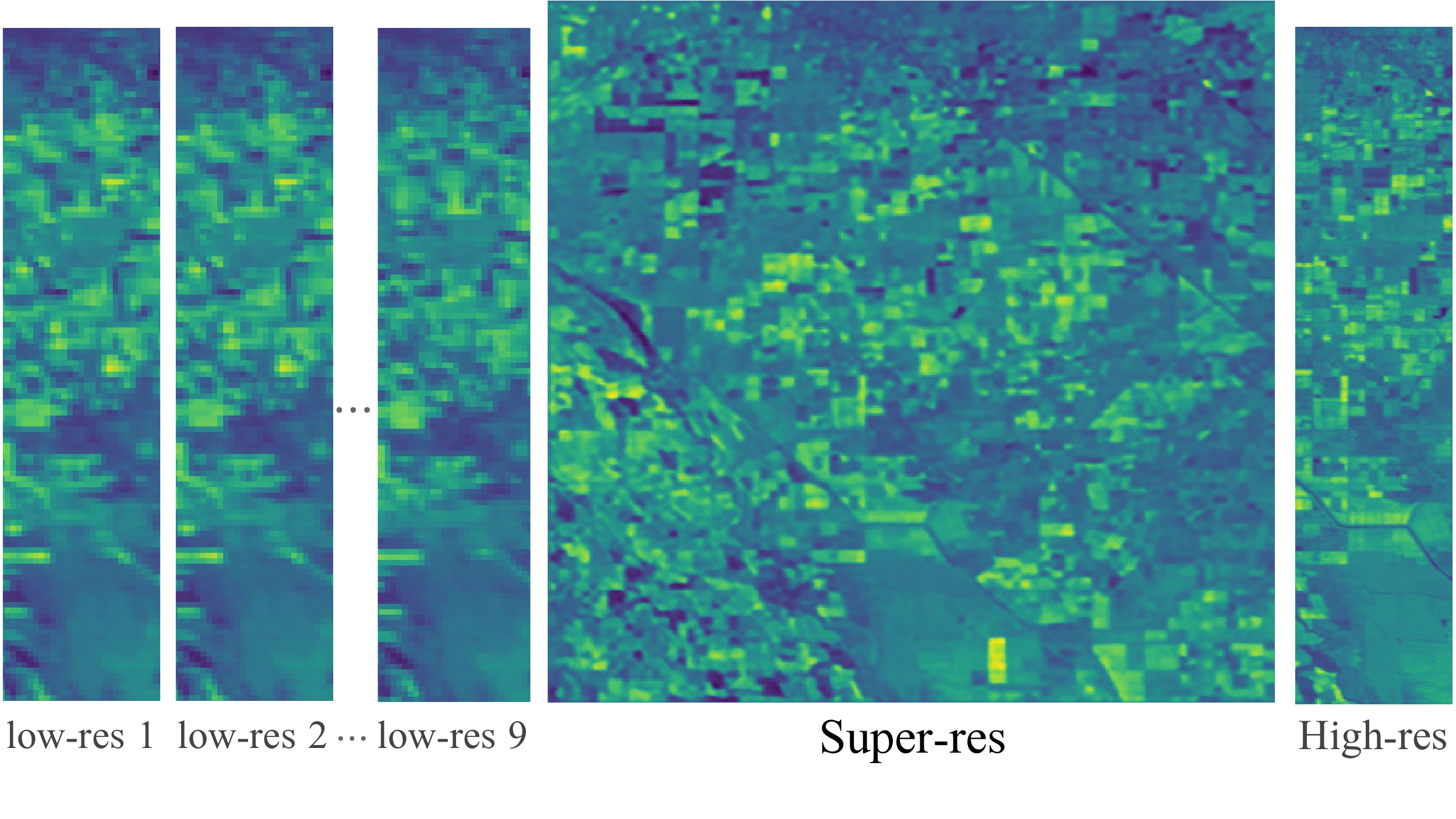}
        \includegraphics[width=0.49\linewidth, trim={0 6ex 0 0}, clip]{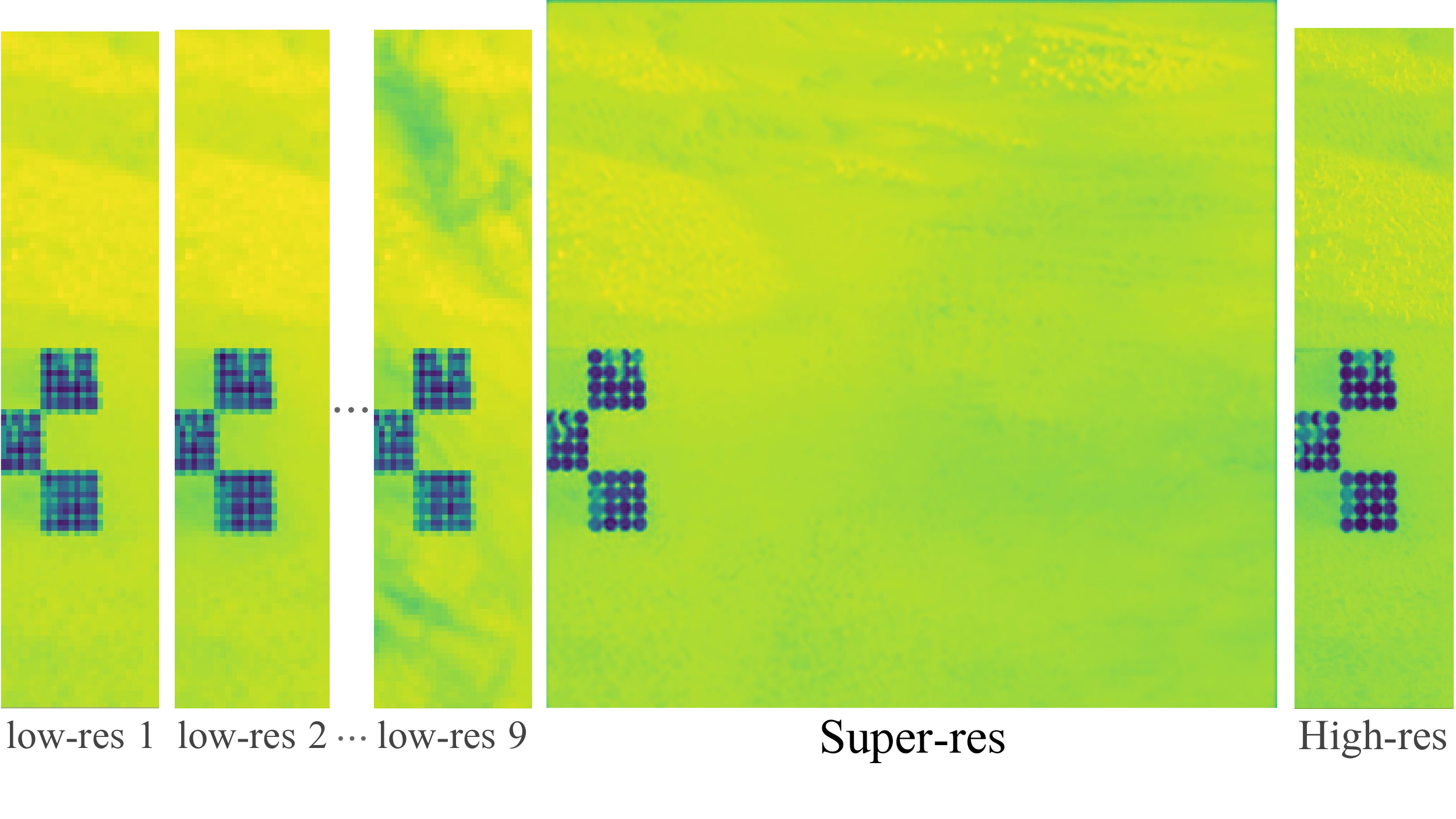}
    \end{center}
    \caption{\footnotesize
        HighRes-net combines many low-resolution images (300 meters/pixel) into one image of superior resolution.
        The same site shot in high-resolution (100m/pix) is also shown for reference.
        Source of low-res and high-res: \texttt{imgset1087} and \texttt{imgset0285} of PROBA-V dataset, see section \ref{sec:5.experiments}.
    }
    \label{fig:Kelvin1}
\end{figure*}

Multiple low-resolution images of the same scene contain collectively more information than any individual low-resolution image, due to minor geometric displacements, e.g. shifts, rotations, atmospheric turbulence, and instrument noise.
Multi-Frame Super-Resolution (MFSR) \citep{tsai1984multiple} aims to reconstruct hidden high-resolution details from multiple low-resolution views of the same scene.
Single Image Super-Resolution (SISR), as a special case of MFSR, has attracted much attention in the computer vision and deep learning communities in the last 5 years, with neural networks learning complex image priors to upsample and interpolate images \citep{xu2014deep, srivastava2015highway, he2016deep}.
However, not much work has explored the learning of representations for the more general setting of MFSR to address the additional challenges of co-registration and fusion of multiple low-resolution images.

This paper shows how Multi-Frame Super-Resolution (MFSR) --- a valuable capability in remote sensing --- can benefit from recent advances in deep learning.
Specifically, this work is the first to introduce a deep-learning approach that solves the co-registration, fusion and registration-at-the-loss problems in an end-to-end learning fashion.

This research is driven by the proliferation of planetary-scale Earth Observation to monitor climate change and the environment.  
Such observation can be used to inform policy, achieve accountability and direct on-the-ground action, e.g. within the framework of the Sustainable Development Goals \citep{Jensen:digitalecosystem}.

\subsection{Nomenclature}

First, we cover some useful terminology:

\textbf{Registration}
is the problem of estimating the relative geometric differences between two images (e.g. due to shifts, rotations, deformations).

\textbf{Fusion}
in the context of MFSR, is the problem of mapping multiple low-resolution representations into a single representation.

\textbf{Co-registration}
is the problem of registering all low-resolution views to improve their fusion.

\textbf{Registration-at-the-loss}
is the problem of registering the super-resolved (SR) reconstruction to the high-resolution (HR) target image prior to computing the loss function.
Similarly, a \textit{registered loss} function is effectively invariant to geometric differences between the reconstruction and the target, whereas a conventional (un-registered) loss would cause undue changes to even a perfect reconstruction.

Co-registration of multiple images is required for longitudinal studies of land change and environmental degradation.
The fusion of multiple images is key to exploiting cheap, high-revisit-frequency satellite imagery, but of low-resolution, moving away from the analysis of infrequent and expensive high-resolution images.
Finally, beyond fusion itself, super-resolved generation is required throughout the technical stack:
both for labeling, but also for human oversight
\citep{drexler2019reframing}
demanded by legal context \citep{drake:2018}.

\subsection{Contributions}

\paragraph{HighRes-net}
We propose a deep architecture that learns to fuse an arbitrary number of low-resolution frames with implicit co-registration through a reference-frame channel. 

\paragraph{ShiftNet}
Inspired by HomographyNet \citep{detone2016deep}, we define a model that learns to register and align the super-resolved output of HighRes-net, using ground-truth high-resolution frames as supervision.
This registration-at-the-loss mechanism enables more accurate feedback from the loss function into the fusion model, when comparing a super-resolved output to a ground truth high resolution image.
Otherwise, a MFSR model would naturally yield blurry outputs to compensate for the lack of registration, to correct for sub-pixel shifts and account for misalignments in the loss.

\paragraph{End-to-end fusion + registration}
By combining the two components above, we propose the first architecture to learn the tasks of fusion and registration in an end-to-end fashion.

We test and compare our approach to several baselines on real-world imagery from the \href{\URLESAPROBAV}{PROBA-V satellite of ESA}.
Our performance has topped the Kelvins competition on MFSR, organized by the Advanced Concepts Team of ESA \citep{mrtens2019superresolution} (see section \ref{sec:5.experiments}).


The rest of the paper is divided as follows:
in Section \ref{sec:2.background}, we discuss related work on SISR and MFSR;
Section \ref{sec:3.hrnet} outlines HighRes-net and section \ref{sec:4.registration} presents ShiftNet, a differentiable registration component that drives our registered loss mechanism during end-to-end training.
We present our results in section \ref{sec:5.experiments}, and in Section \ref{sec:6.discussion} we discuss some opportunities for and limitations and risks of super-resolution.


\section{Background}
\label{sec:2.background}

\begin{figure}[ht]
    \centering
    \includegraphics[width=.99\linewidth]{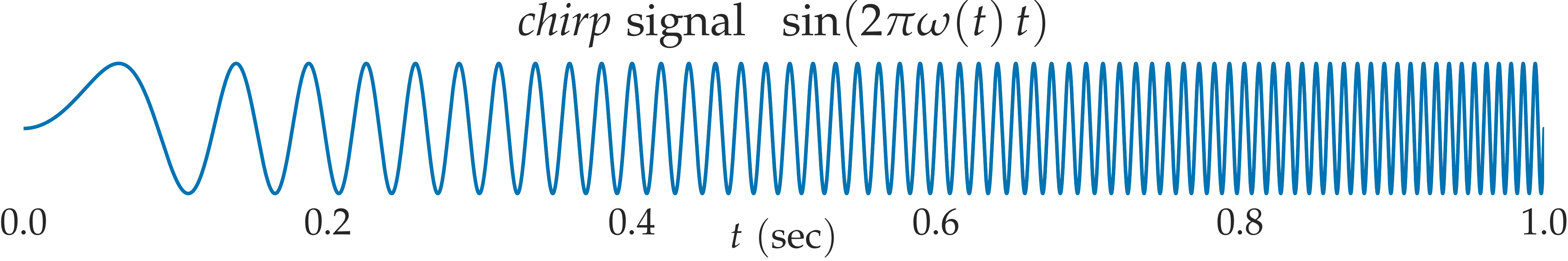}\\
    \includegraphics[width=.99\linewidth]{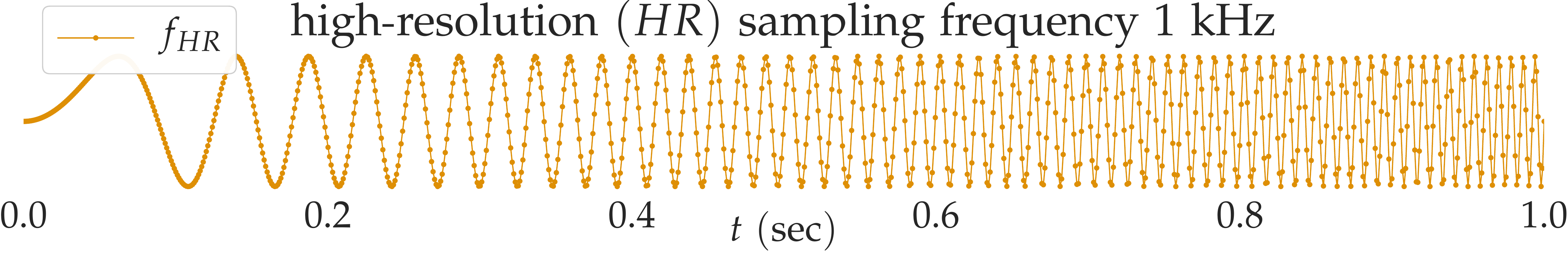}
    \hspace{2mm}\includegraphics[width=.99\linewidth]{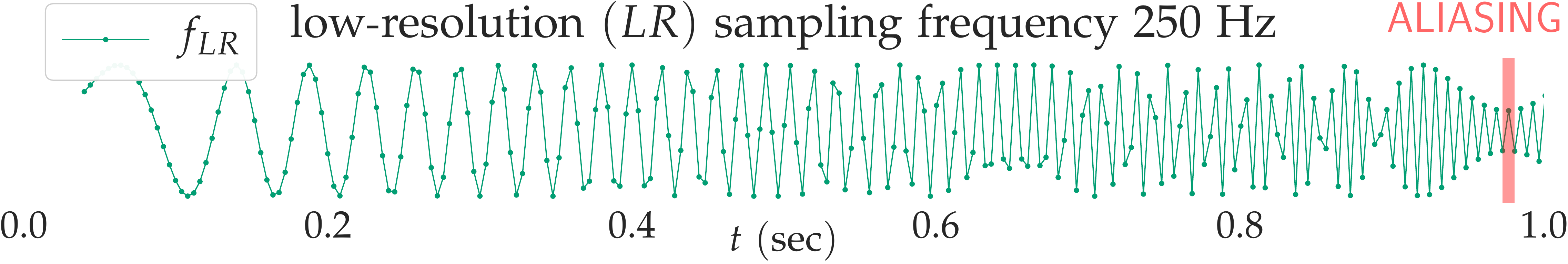}
    \caption{
        \small
        An example of \textit{aliasing} (shown with red at its most extreme).
        \textbf{Top}:
        A \textit{chirp} harmonic oscillator $\sin\paren{2 \pi \omega(t) t}$, with instantaneous frequency $\omega(t)$.
        \textbf{Left}:
        The shape of the high-resolution sample resembles the underlying chirp signal.
        \textbf{Right}:
        Close to $t=1$, the apparent frequency of the low-resolution sample does not match that of the chirp.
        It happens when the sampling rate falls below the \textit{Nyquist rate},
        $2~s$, where $s$
        is the highest non-zero frequency of the signal.
    }
    \label{fig:1D_sim}
\end{figure}

How much detail can we resolve in the digital sample of some natural phenomenon?
\cite{nyquist1928certain} observed that it depends on the instrument's sampling rate and the oscillation frequency of the underlying natural signal.
\cite{shannon1949communication} built a sampling theory that explained Nyquist's observations when the sampling rate is constant (\textit{uniform} sampling) and determined the conditions of \textit{aliasing} in a sample.
Figure \ref{fig:1D_sim} illustrates this phenomenon.

Sampling at high-resolution (left) maintains the frequency of the chirp signal (top).
Sampling at a lower resolution (right), this apparent chirped frequency is lost due to aliasing, which means that the lower-resolution sample has a fundamentally smaller capacity for resolving the information of the natural signal, and a higher sampling rate can resolve more information.

Shannon's sampling theory has since been generalized for multiple interleaved sampling frames \citep{papoulis1977generalized, marks2012introduction}.
One result of the generalized sampling theory is that we can go beyond the Nyquist limit of any individual uniform sample by interleaving several uniform samples taken concurrently. 
When an image is down-sampled to a lower resolution, its high-frequency details are lost permanently and cannot be recovered from any 
image in isolation.
However, by combining multiple low-resolution images, it is possible to recover the original scene at a higher resolution.

\subsection{Multi-frame / multi-image super-resolution}


Different low-resolution samples may be sampled at different phase shifts, such that the same high-resolution frequency information will be packed with various phase shifts.
Consequently, when multiple low-resolution samples are available, the fundamental problem of MFSR is one of fusion by \textit{de-aliasing} --- that is, to disentangle the high-frequency components packed in low-resolution imagery.

Such was the first work on MSFR by \citet{tsai1984multiple}, who framed the reconstruction of a high-resolution image by fusion of low-resolution images in the Fourier domain, assuming that their phase shifts are known.
However, in practice the shifts are never known, therefore the fusion problem must be tackled in conjunction with the \textit{registration} problem \citep{irani1991improving, fitzpatrick2000image, capel2001super}.
Done the right way, a composite \emph{super-resolved} image can reveal some of the original high-frequency detail that would have been unrecoverable from a single low-resolution image.

Until now, these tasks have been learned and / or performed separately.
So any incompatible inductive biases that are not reconciled through co-adaptation, would limit the applicability of that particular fuser-register combination.
To that end, we introduce HighRes-Net, the first fully end-to-end deep architecture for MFSR settings, that jointly learns and co-adapts the fusion and (co-)registration tasks to one another.

\subsubsection{Video and Stereo super-resolution}

The setting of MFSR is related to Video SR and Stereo SR, although MFSR is less constrained in a few important ways:
In MFSR, a model must fuse and super-resolve from sets, not sequences, of low-resolution inputs.
The training input to our model is an unordered set of low-resolution views, with unknown timestamps.
The target output is a single high-resolution image --- not another high-resolution video or sequence.
When taken at different times, the low-resolution views are also referred to as \textit{multi-temporal} (see e.g. \citealt{molini2019deepsum}).

In Video SR, the training input is a temporal sequence of frames that have been synthetically downscaled.
An auto-regressive model (one that predicts at time $t = T$ based on past predictions at $t < T$), benefits from this additional structure by estimating the motion or optical flow in the sequence of frames \citep{tao2017detail, sajjadi2018frame, yan2019frame, wang2019edvr}.

In Stereo SR, the training input is a pair of low-resolution images shot simultaneously with a stereoscopic camera, and the target is the same pair in the original high-resolution.
The problem is given additional structure in the prior knowledge that the pair differ mainly by a parallax effect
\citep{wang2019deep, wang2019learning}.

\begin{figure*}[ht]
    \centering
    \subfigure[HighRes-net]{
        \includegraphics[width=0.53\linewidth, trim={0 13ex 45ex 9ex}, clip]{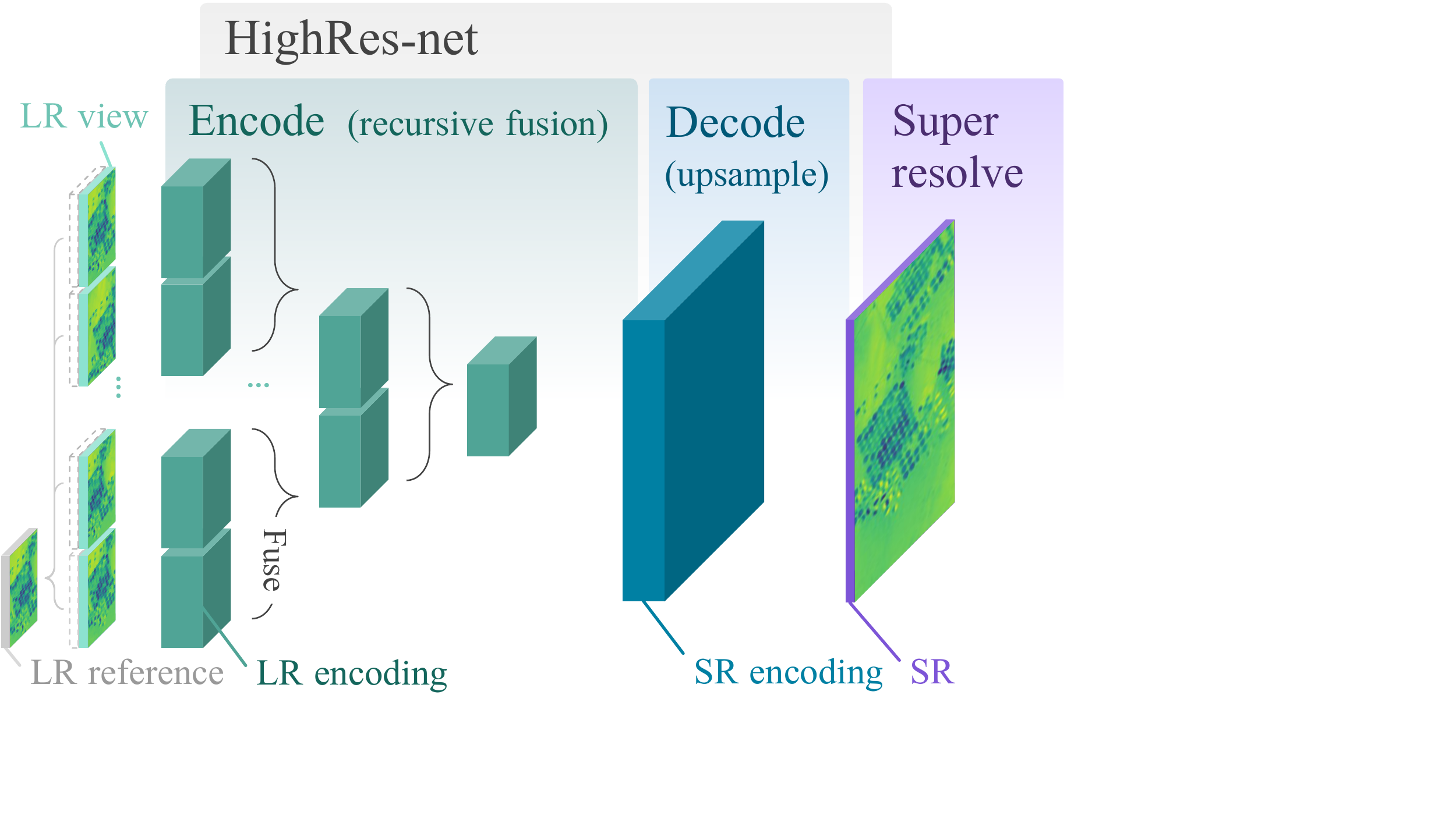}
    	\label{fig:hrnet}
    }
    \hspace*{\fill}
    \subfigure[Registered loss]{
        \includegraphics[width=.42\linewidth, trim={0 30ex 75ex 0}, clip]{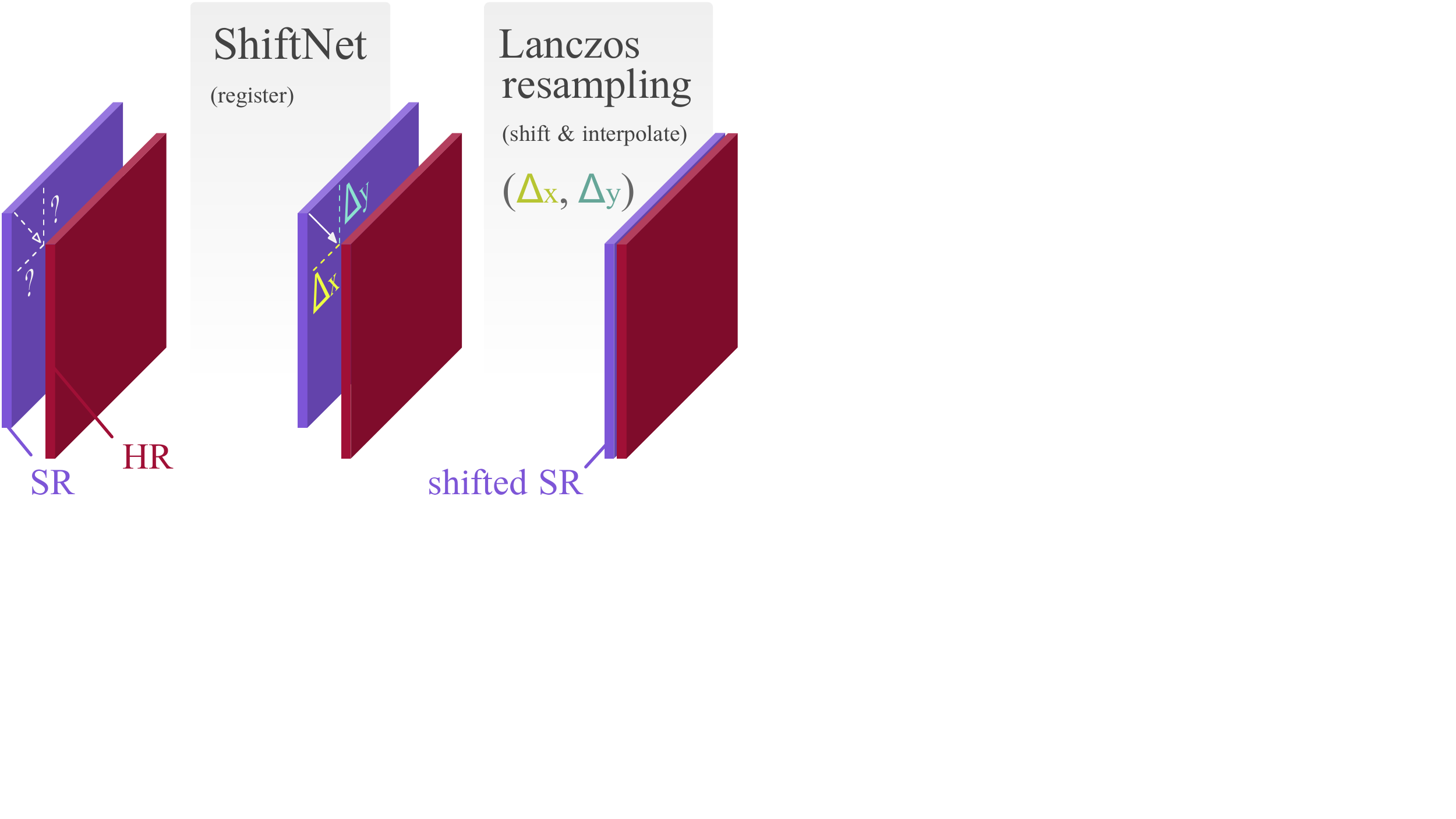}
    	\label{fig:shiftnet_lanczos}
    }
    \caption{ \label{fig:archs}
        \footnotesize
        Schematic of the full processing pipeline, trained end-to-end. At test time, only HighRes-net is used.
        \subref{fig:hrnet}
        \textbf{HighRes-net}: In the \textit{Encode} stage, an arbitrary number of LR views are paired with the \textit{reference} low-resolution image (the median low-resolution in this work).
        Each LR view--reference pair is encoded into a \textit{view-specific} latent representation.
        The LR encodings are fused recursively into a single \textit{global} encoding.
        In the \textit{Decode} stage, the global representation is upsampled by a certain zoom factor ($\times 3$ in this work).
        Finally, the \textit{super-resolved} image is reconstructed by combining all channels of the upsampled global encoding.
        \subref{fig:shiftnet_lanczos}
        \textbf{Registered loss}:
        Generally, the reconstructed SR will be shifted with respect to the ground-truth HR.
        ShiftNet learns to estimate the $\paren{\Delta x, \Delta y}$ shift that improves the loss.
        Lanczos resampling: $\paren{\Delta x, \Delta y}$ define two $1$D shifting Lanczos kernels that translate the SR by a separable convolution.
    }
\end{figure*}

\subsection{Generative perspective}

In addition to aliasing, MFSR deals with random processes like noise, blur, geometric distortions -- all contributing to random low-resolution images.
Traditionally, MFSR methods have assumed prior knowledge of the motion model, blur kernel, noise and degradation process that generate the data; see for example,  \cite{pickup2006optimizing}.
Given multiple low-resolution images, the challenge of MFSR is to reconstruct a plausible image of higher-resolution that could have generated the observed low-resolution images.
Optimization methods aim to improve an initial guess by minimizing an error between simulated and observed low-resolution images. 
These methods traditionally model the additive noise $\eps$ and prior knowledge about natural images explicitly, to constrain the parameter search space and derive objective functions, using e.g. Total Variation \citep{chan1998total, farsiu2004fast}, Tikhonov regularization \citep{nguyen2001computationally} or Huber potential \citep{pickup2006optimizing} to define appropriate constraints on images.



In some situations, the image degradation process is complex or not available, motivating the development of nonparametric strategies.
Patch-based methods learn to form high-resolution images directly from low-resolution patches, e.g. with k-nearest neighbor search \citep{freeman2002example, chang2004super}, sparse coding and sparse dictionary methods \citep{yang2010image, zeyde2010single, kim2010single}).
The latter represents images in an over-complete basis and allows for sharing a prior across multiple sites.

In this work, we are particularly interested in super-resolving satellite imagery.
Much of the recent work in Super-Resolution has focused on SISR for natural images.
For instance, \cite{dong2014learning} showed that training a CNN for super-resolution is equivalent to sparse coding and dictionary based approaches.
\cite{kim2016deeply} proposed an approach to SISR using recursion to increase the receptive field of a model while maintaining capacity by sharing weights.
Many more networks and learning strategies have recently been introduced for SISR and image deblurring. 
Benchmarks for SISR \citep{Timofte_2018_CVPR_Workshops}, differ mainly in their upscaling method, network design, learning strategies, etc.
We refer the reader to \citep{wang2019deep} for a more comprehensive review.

Few deep-learning approaches have considered the more general MFSR setting and attempted to address it in an end-to-end learning framework.
Recently, \cite{kawulok2019deep} proposed a \emph{shift-and-add} method and suggested ``including image registration" in the learning process as future work.

In the following sections, we describe our approach to solving both aspects of the registration problem -- co-registration and registration-at-the-loss -- in a memory-efficient manner.


\section{HighRes-net: MFSR by recursive fusion}
\label{sec:3.hrnet}

In this section, we present HighRes-net, a neural network  for multi-frame super-resolution inside a single spectral band (greyscale images), using joint co-registration and fusion of multiple low-resolution views in an end-to-end learning framework.
From a high-level, HighRes-net consists of an encoder-decoder architecture and can be trained by stochastic gradient descent using high-resolution ground truth as supervision, as shown in Figure \ref{fig:archs}.

\paragraph{Notation}
We denote by $\theta$ the parameters of HighRes-net trained for a given upscaling factor $\gamma$.
$LR_{v,i} \in \mathbb{R}^{C \times W \times H}$ is one of a set of $K$ low-resolution views from the same site $v$, where $C$, $W$ and $H$ are the number of input channels, width and height of $LR_{v,i}$, respectively.
We denote by $SR^{\theta}_v = F_{\theta}^{\gamma}\paren{LR_{v,1}, ~ \dots ~ , ~ LR_{v,K}}$, the output of HighRes-net and by $HR_v \in \mathbb{R}^{C \times \gamma W \times \gamma H}$ a ground truth high-resolution image.
We denote by $\left[T_{1}, T_{2}\right]$ the concatenation of two images channel-wise.
In the following we supress the index $v$ over sites for clarity.

HighRes-Net consists of three main steps: (1) encoding, which learns relevant features associated with each low-resolution view, (2) fusion, which merges relevant information from views within the same scene, and (3) decoding, which proposes a high-resolution reconstruction from the fused summary.

\subsection{Encode, Fuse, Decode}
\label{subsec:encode_fude_decode}

\paragraph{Embed, Encode}
The core assumption of MFSR is that the low-resolution image set contains collectively more information than any single low-resolution image alone, due to differences in photometric or spatial coverage for instance.
However, the redundant low frequency information in multiple views can hinder the training and test performance of a MFSR model.
We thus compute a reference image {\em ref} as a \emph{shared representation} for multiple low-resolution views $\left(LR_{i}\right)_{i = 1}^{K}$ and embed each image jointly with {\em ref}.
This highlights differences across the multiple views  \citep{sanchez2019learning},
and potentially allows HighRes-net to focus on difficult high-frequency features such as crop boundaries and rivers during super-resolution.
The \emph{shared} representation or \textit{reference} image intuitively serves as an anchor for implicitly aligning and denoising multiple views in deeper layers.
We refer to this mechanism as \textit{implicit co-registration}.

HighRes-net's embedding layer $\text{emb}_{\theta}$ consists of a convolutional layer and two residual blocks with PReLu activations \citep{He:2015:DDR:2919332.2919814} and is shared across all views. 
The embedded hidden states $s^{0}_{i}$ are computed in parallel as follows:
\begin{align}
    \hspace{-2ex} \mathrm{\textit{ref}}~(c,\!i,\!j)
    &= \mathrm{median}\paren{LR_{1}(c,\!i,\!j), \dots, LR_{K}(c,\!i,\!j)},
    \\
    s^{0}_{i} &= \text{emb}_{\theta} \paren{\left[LR_{i}, ~ \mathrm{\textit{ref}} ~\right]} \in \mathbb{R}^{C_{h} \times W \times H},
\end{align}
where $\mathrm{\textit{ref}} \in \mathbb{R}^{C \times W \times H}$, and $C_h$ denotes the channels of the hidden state.

The imageset is padded if the number of low-resolution views $K'$ is not a power of 2:
we pad the set with dummy zero-valued views, such that the new size of the imageset $K$ is the next power of 2.
See Algorithm \ref{algo:HRnet_forward}, line 1.

\paragraph{Fuse}

\begin{figure}[h]
\begin{center}
    \includegraphics[width=1.00\linewidth, trim={10ex 135ex 40ex 60ex}, clip]{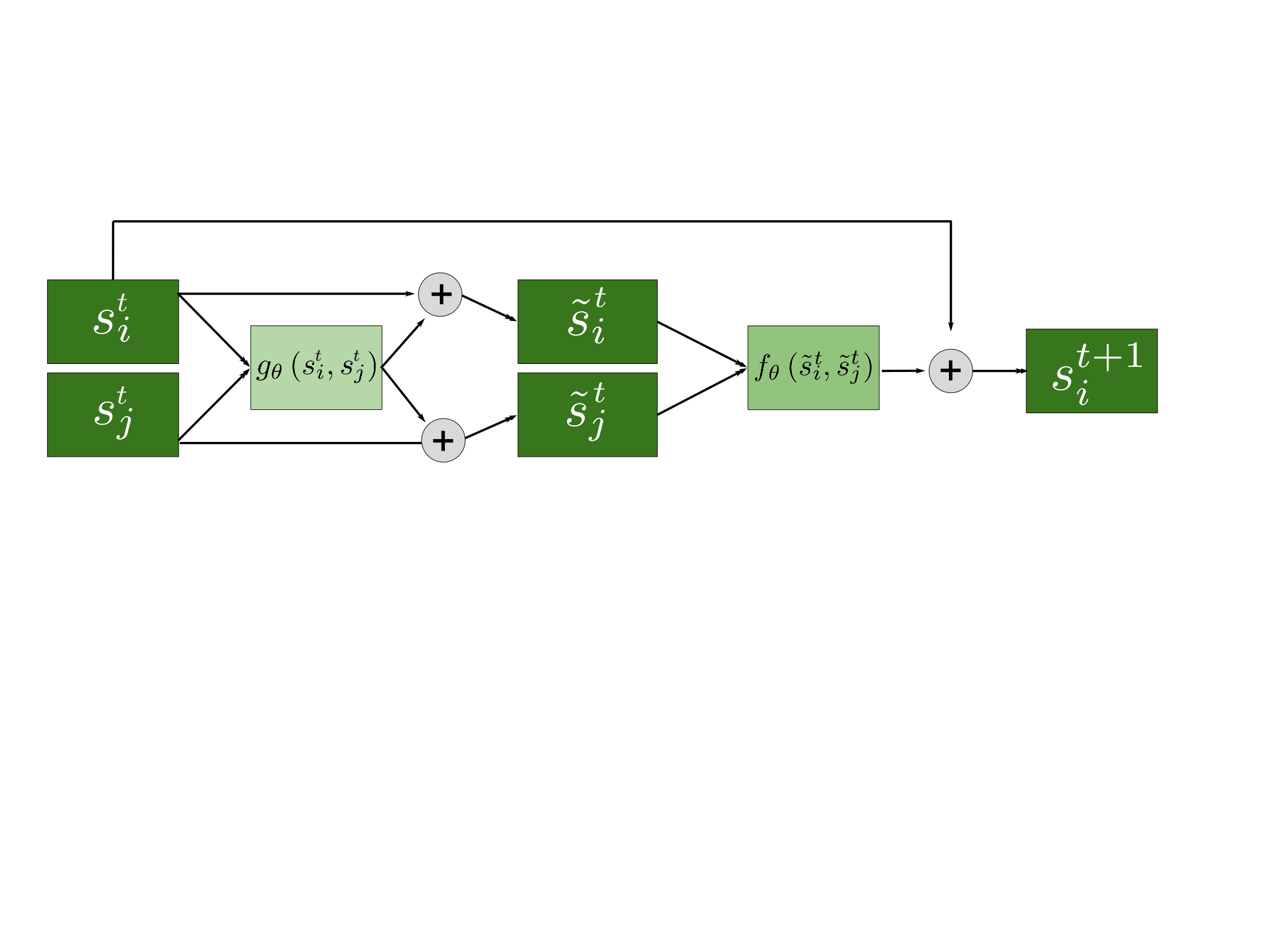}
\end{center}
    \caption{
        \footnotesize
        HighRes-net's global fusion operator consists of a \emph{co-registration} $g_{\theta}$ and a \emph{fusion} $f_{\theta}$ block which aligns and combines two representations into a single representation.
    }
    \label{fig:alpha}
\end{figure}

The embedded hidden states $s^{0}_{i}$ are then fused recursively, halving by two the number of low-resolution states at each fusion step $t$, as shown in Figure~\ref{fig:alpha}.
Given a pair of hidden states $s^{t}_{i}, s^{t}_{j}$, HighRes-net computes a new representation:
\begin{align}
    \label{eq:inner_skip}
    \left[ \tilde{s}^{t}_{i}, ~ \tilde{s}^{t}_{j} \right] 
        &= \left[ s^{t}_{i}, ~ s^{t}_{j} \right] 
        + g_{\theta} \paren{ \left[ s^{t}_{i}, ~ s^{t}_{j} \right] } \in \mathbb{R}^{2C_{h} \times W \times H} \\
    \label{eq:outer_skip}
    s^{t+1}_{i} &= s^{t}_{i} 
        + \alpha_{j} f_{\theta} \paren{ \tilde{s}^{t}_{i}, ~ \tilde{s}^{t}_{j} } \in \mathbb{R}^{C_{h} \times W \times H},
\end{align}
\noindent where $\tilde{s}^{t}_{i}, ~ \tilde{s}^{t}_{j}$ are intermediate representations;
$g_{\theta}$ is a shared-representation within an inner residual block (\eqref{eq:inner_skip});
$f_{\theta}$ is a fusion block,
and $\alpha_{j}$ is 0 if the $j$-th low-resolution view is part of the padding, and 1 otherwise.
$f_{\theta}$ squashes $2C_{h}$ input channels into $C_{h}$ channels and consists of a (conv2d+PreLu).
Intuitively, $g_{\theta}$ \textit{aligns} the two representations and it consists of two (conv2d + PreLU) layers.

The blocks $(f_{\theta}, g_{\theta})$ are shared across all pairs and depths, giving it the flexibility to deal with variable size inputs and significantly reduce the number of parameters to learn.

\paragraph{Upscale and Decode}
After $T = \log_{2}K$ fusion layers, the final low-resolution encoded state $s^{T}_{i}$ contains information from all $K$ input views.
Any information of a spatial location that was initially missing from $LR_i$, is now encoded implicitly in $s^{T}_{i}$.
$T$ is called the depth of HighRes-net.
Only then, $s^{T}_{i}$ is upsampled with a deconvolutional layer \citep{xu2014deep} to a higher-resolution space $s^{T}_{HR} \in \mathbb{R}^{C_{h}  \times \gamma W  \times \gamma H}$.
The hidden high-resolution encoded state $s^{T}_{HR}$ is eventually convolved with a $1 \! \times \! 1$ 2D kernel to produce a final super-resolved image $SR^{\theta} \in \mathbb{R}^{C \times \gamma W  \times \gamma H}$.

The overall architecture of HighRes-net is summarized in Figure \ref{fig:hrnet} and the pseudo-code for the forward pass is given in Algorithm \ref{algo:HRnet_forward}.

\begin{algorithm}
\caption{HighRes-net forward pass}
\label{algo:HRnet_forward}
\begin{algorithmic}
    \vspace{.5ex}
    \STATE {\bfseries Input:} low-resolution views $LR_{1} \dots LR_{K'}$
    \STATE \texttt{\scriptsize \# pad inputs to next power of 2}
    \STATE $\paren{LR_{1} \dots LR_{K}, ~ \alpha_{1}  \dots \alpha_{K}} ~ = ~ \mathrm{pad}\paren{LR_{1}\dots LR_{K'}}$
    \STATE $s^{0}_{i} ~ = ~ \mathrm{\color{red} encode}\paren{LR_{i}}$ \texttt{\scriptsize // parallelized across $K$ views}
    \STATE $T ~ = ~ \log_{2} K$ 
    \STATE $k ~ = ~ K$
    \FOR{$t ~ = ~ 1 \ldots T$}
        \FOR{$i ~ = ~ 1 \ldots k/2$}
            \STATE \hspace{-1ex} $s^{t}_{i} ~ = ~ \mathrm{\color{red} fuse}\paren{s^{t-1}_{i}, ~ s^{t-1}_{k-i}, ~ \alpha_{k-i}}$ \texttt{\scriptsize \# fuse encodings}
        \ENDFOR
        \STATE $k = k/2$
    \ENDFOR
    \STATE $SR ~ = ~ \mathrm{\color{red} decode}\paren{s^{T}_i}$ \texttt{\scriptsize \# output super-resolved view}
\end{algorithmic}
\end{algorithm}


\section{Registration matters}
\label{sec:4.registration}


Co-registration matters for fusion.
HighRes-net learns to implicitly co-register multiple low-resolution views $LR_{i}$ and fuse them into a single super resolved image $SR_{\theta}$.
We note that since the recursive fusion stage accepts only the encoded low-resolution / reference pairs.
So no aspect of our low-res co-registration scheme comes with the built-in assumption that the difference in low-resolution images must be explained only by translational motion.

A more explicit registration-at-the-loss can also be used for measuring similarity metrics and distances between $SR_{\theta}$ and $HR$.
Indeed, training HighRes-Net alone, by minimizing a reconstruction error such as the mean-squared error  between $SR_{\theta}$ and $HR$, leads to blurry outputs, since the neural network has to compensate for pixel and sub-pixel misalignments between its output $SR_{\theta}$ and $HR$.

Here, we present ShiftNet-Lanczos, a neural network that can be paired with HighRes-net to account for pixel and sub-pixel shifts in the loss, as depicted in Figure \ref{fig:shiftnet_lanczos}.
Our ablation study A.2 and qualitative visual analysis suggest that this strategy helps HighRes-net learn to super-resolve and leads to clearly improved results.

\subsection{ShiftNet-Lanczos}


ShiftNet learns to align a pair of images with sub-pixel translations.
ShiftNet registers pairs of images by predicting two parameters defining a global translation.
Once a sub-pixel translation is found for a given pair of images, it is applied through a Lanczos shift kernel to align the images.

\paragraph{ShiftNet}
The architecture of ShiftNet is adapted from HomographyNet \citep{detone2016deep}.
Translations are a special case of homographies.
In this sense, ShiftNet is simply a special case of HomographyNet, predicting
2 shift parameters instead of 8 homography parameters.
See Supplementary Material \ref{appendix:registration_matters}, for details on the architecture of ShiftNet.

One major difference from HomographyNet is the way we train ShiftNet:
In \citep{detone2016deep}, HomographyNet is trained on synthetically transformed data, supervised with ground-truth homography matrices.
In our setting, ShiftNet is trained to cooperate with HighRes-net, towards the common goal of MFSR (see section \textbf{Objective function} below).

\paragraph{Lanczos kernel for shift / interpolation}
To shift and align an image by a sub-pixel amount, it must be convolved with a filter that shifts for the integer parts and interpolates for the fractional parts of the translation.
Standard options for interpolation include the nearest-neighbor, sinc, bilinear, bicubic, and Lanczos filters \citep{turkowski1990filters}.
The sinc filter has an infinite support as opposed to any digital signal, so in practice it produces ringing or ripple artifacts --- an example of the Gibbs phenomenon.
The nearest-neighbor and bilinear filters do not induce ringing, but strongly attenuate the higher-frequency components (over-smoothing), and can even alias the image.
The Lanczos filter reduces the ringing significantly by using only a finite part of the sinc filter (up to a few lobes from the origin).
Experimentally, we found the Lanczos filter to perform the best.

\subsection{Objective function}

In our end-to-end setting, registration improves super-resolution as HighRes-net receives more informative gradient signals when its output is aligned with the ground truth high-resolution image.
Conversely, super-resolution benefits registration, since good features are key to align images \citep{clement2018image}.
We thus trained HighRes-Net and ShiftNet-Lanczos in a cooperative setting, where both neural networks work together to minimize an objective function, as opposed to an adversarial setting where a generator tries to fool a discriminator.
HighRes-net infers a latent super-resolved variable and ShiftNet maximizes its similarity to a ground truth high-resolution image with sub-pixel shifts.

By predicting and applying sub-pixel translations in a differentiable way, our approach for registration and super-resolution can be combined in an end-to-end learning framework.
Shift-Net predicts a sub-pixel shift $\Delta$ from a pair of high-resolution images.
The predicted transformation is applied with Lanczos interpolation to align the two images at a pixel level.
ShiftNet and HighRes-Net are trained end-to-end, to minimize a joint loss function.
Our objective function is composed of a registered reconstruction loss, described in Algorithm \ref{algo:ShiftNet_forward}.

\begin{algorithm}
\caption{Sub-pixel registered loss} 
\label{algo:ShiftNet_forward}
\begin{algorithmic}
    \vspace{.5ex}
    \STATE \hspace{-2ex} {\bfseries Input:} super-resolution $SR_{\theta}$, ground-truth high-res $HR$
    \vspace{.5ex}
    \STATE \hspace{-2ex} $\paren{\Delta x, \Delta y} = \mathrm{ShiftNet}\paren{SR_{\theta}, HR}$ \texttt{\scriptsize \# register $SR$ to $HR$}
    \vspace{.5ex}
    \STATE \hspace{-2ex} \texttt{\scriptsize \# 1D Lanczos kernels for $x$ and $y$ sub-pixel shifts}
    \STATE \hspace{-2ex} $\paren{\kappa_{\Delta x}, ~ \kappa_{\Delta y}} = \mathrm{LanczosShiftKernel}\paren{\Delta x, ~ \Delta y}$
    \vspace{.5ex}
    \STATE \hspace{-2ex} \texttt{\scriptsize \# 2D sub-pixel shift by separable 1D convolutions}
    \STATE \hspace{-2ex} $SR_{\theta,\Delta} ~ = ~ SR_{\theta} ~ * ~ \kappa_{\Delta x} ~ * ~ \kappa_{\Delta y}$
    \vspace{.5ex}
    \STATE \hspace{-2ex} \texttt{\scriptsize \# sub-pixel registered loss}
    \STATE \hspace{-2ex} $\ell_{\theta,\Delta} ~ = ~ \mathrm{loss} \paren{SR_{\theta,\Delta}, ~ HR }$  \texttt{\scriptsize \# $\ell_{\theta,\Delta} \leqslant \mathrm{loss} \paren{SR_{\theta}, HR}$}
\end{algorithmic}
\end{algorithm}

\paragraph{Leaderboard score \& loss}
We iterated our design based on our performance on the leaderboard of the ESA competition.
The leaderboard is ranked by the cPSNR (clear Peak Signal-to-Noise Ratio) score.
It is similar to the mean-squared error, but also corrects for brightness bias and clouds in satellite images \citep{mrtens2019superresolution}, but the proposed architecture is decoupled from the choice of loss.


See Algorithm~\ref{algo:ShiftNet_forward} for computing the registered loss $\ell_{\theta,\Delta}$.
We further regularize the L2 norm of ShiftNet's output with a hyperparameter $\lambda$ and our final joint objective is given by:
\begin{equation}
    L_{\theta,\Delta}(SR_{\theta},HR) = \ell_{\theta,\Delta} + \lambda ||\Delta||_{2}
\end{equation}


\section{Experiments}
\label{sec:5.experiments}

The PROBA-V satellite carries two different cameras for capturing a high-resolution / low-resolution pair.
This makes the PROBA-V dataset one of the first publicly available datasets for MFSR with naturally occurring low-resolution and high-resolution pairs of satellite imagery.

\paragraph{Perils of synthetic data}
This is in contrast to most of the work in SR (Video, Stereo, Single-Image, Multi-Frame), where it is common practice to train with low-resolution images that have been artificially generated through simple bilinear down-sampling
\footnote{See also, image restoration track at CVPR19. The vast majority of challenges were performed on synthetically downscaled / degraded images: REDS4: ``NTIRE 2019 challenge on video deblurring: Methods and results", \href{http://www.vision.ee.ethz.ch/ntire19}{NTIRE Workshop at CVPR2019}.}
\citep{bulat2018learn, wang2019flickr1024, nah2019ntire}.
Methods trained on artificially downscaled datasets are biased, in the sense that they learn to undo the action of a simplistic downscaling operator.
On the downside, they also tune to its inductive biases.
For example, standard downsampling kernels (e.g. bilinear, bicubic) are simple low-pass filters, hence the eventual model implicitly learns that all input images come from the same band-limited distribution.
In reality, natural complex images are only approximately band-limited \citep{ruderman1994statistics}.
Methods that are trained on artificially downscaled datasets fail to generalize to real-world low-resolution, low quality images \citep{shocher2018zero}.
For this reason, we experiment only on PROBA-V, a dataset that does not suffer from biases induced by artificial down-sampling.

\subsection{Proba-V Kelvin dataset}

The performance of our method is illustrated with satellite imagery from the Kelvin competition, organized by ESA's Advanced Concept Team (ACT).

The Proba-V Kelvin dataset \citep{mrtens2019superresolution} contains 1450 scenes (RED and NIR spectral bands) from 74 hand-selected Earth regions around the globe at different points in time.
The scenes are split into 1160 scenes for training and 290 scenes for testing.
Each data-point consists of exactly one 100m resolution image as $384 \times 384$ greyscale pixel images (HR) and several 300m resolution images from the same scene as $128 \times 128$ greyscale pixel images (LR), spaced days apart.
We refer the reader to the Proba-V manual \citep{wolters2014proba} for further details on image acquisition.

Each scene comes with at least 9 low-resolution views, and an average of 19.
Each view comes with a noisy \textit{quality map}.
The quality map is a binary map, that indicates concealed pixels due to volatile features, such as clouds, cloud shadows, ice, water and snow.
The sum of clear pixels (1s in the binary mask) is defined as the \textit{clearance} of a low-resolution view.
These incidental and noisy features can change fundamental aspects of the image, such as the contrast, brightness, illumination and landscape features.
We use the \textit{clearance} scores to randomly sample from the imageset of low-resolution views, such that views with higher \textit{clearance} are more likely to be selected.
This strategy helps to prevent overfitting.
See Supplementary Material \ref{appendix:permutation_invariance} for more details.

\paragraph{Working with missing \& noisy values}
A quality map can be used as a \textit{binary mask} to indicate noisy or occluded pixels, due to clouds, snow, or other volatile objects.
Such a mask can be fed as an additional input channel in the respective low-resolution view, in the same fashion as the reference frame.
When missing value masks are available, neural networks can learn which parts of the input are anomalous, noisy, or missing, when provided with such binary masks (see e.g. \cite{che2018recurrent}).
In satellite applications where clouds masks are not available, other segmentation methods would be in order to infer such masks as a preprocessing step (e.g. \cite{long2015fully}).
In the case of the PROBA-V dataset, we get improved results when we make no use of the masks provided.
Instead we use the masks only to inform the sampling scheme within the low-resolution imageset to prevent overfitting.

\subsection{Comparisons}

All experiments use the same hyperparameters, see Supplementary Material \ref{appendix:a}.
By default, each imageset is padded to $32$ views for training and testing, unless specified otherwise. 
Our PyTorch implementation takes less than 9h to train on a single NVIDIA V100 GPU.
At test time, it takes less than 0.2 seconds to super-resolve ($\times 3$ upscaling) a scene with 32 low-resolution $128 \times 128$ views.
Our implementation is available on GitHub
\footnote{\url{https://github.com/ElementAI/HighRes-net}}
.

We evaluated different models on
\href{https://kelvins.esa.int/proba-v-super-resolution}{ESA's Kelvin competition}.
Our best model, HighRes-Net trained jointly with ShiftNet-Lanczos, scored consistently at the top of the public and final leaderboard, see Table \ref{tab:scores}.

We compared HighRes-net to several other approaches:

\textbf{ESA baseline} --- upsamples and averages the subset of low-resolution views with the highest clearance in the set.

\textbf{SRResNet} --- SISR approach by \citet{ledig2017photo}.

\textbf{SRResNet \!+\! ShiftNet} --- trained jointly with ShiftNet.

\textbf{SRResNet-6 +\! ShiftNet} ---
at test time, it independently upsamples 6 low-resolution views, then co-registers/aligns them with \textbf{ShiftNet}, and averages them.

\textbf{ACT} (Advanced Concepts Team, \citealp{mrtens2019superresolution}) --- CNN with 5 channels for the 5 clearest low-resolution views.

\textbf{DeepSUM} \cite{molini2019deepsum} --- like SRResNet-6 \!+\! ShiftNet, it
upsamples independently a fixed number of low-resolution views, co-registers and fuses them.
Here, the co-registration task is learned separately.
One caveat with \textit{upsampling-first} approaches is that the memory cost and training cycle grows quadratically with the upscaling factor.
On the ESA dataset, this means that \textbf{DeepSUM} must train with $3 \times 3$ times the volume of intermediate representations, which can take from several days to a week to train on a NVIDIA V100 GPU.

\textbf{HighRes-net} (trained jointly with ShiftNet, see sections \ref{sec:3.hrnet} and \ref{sec:4.registration}) --- in contrast to \textbf{DeepSUM}, our approach upsamples after fusion, conserving memory, and takes up to 9 hours to train on a NVIDIA V100 GPU.

\textbf{HighRes-net+} --- averages the outputs of two pre-trained
HighRes-net models at test time, one with $K$ bounded to 16 input views and the other to $32$.

\subsection{ESA Kelvin leaderboard}

The Kelvin competition used the \textit{corrected clear} PSNR (cPSNR) quality metric as the standardized measure of performance.
The cPSNR is a variant of the Peak Signal to Noise Ratio (PSNR) used to compensate for pixel-shifts and brightness bias. 
We refer the reader to
\citep{mrtens2019superresolution} for the motivation and derivation of this quality metric.
The cPSNR  metric is normalized by the score of the ESA baseline algorithm so that a score smaller than 1 means ``better than the ESA baseline''.
We also use it as our training objective with sub-pixel registration (see also section \ref{fig:shiftnet_lanczos} on ShiftNet).

\begin{table}[t]
\caption{Public \& final leaderboard cPSNR scores in \href{https://kelvins.esa.int/proba-v-super-resolution}{ESA's Kelvin competition}.
Lower is better.}
\label{tab:scores}
\begin{center}
\begin{small}
\begin{sc}
\begin{tabular}{lcc}
\toprule
Method & Public & Final \\
\midrule
SRResNet & 1.0095 & 1.0084 \\
ESA baseline & 1.0000 & 1.0000 \\
SRResNet \!+\! ShiftNet & 1.0002 & 0.9995 \\
ACT & 0.9874 & 0.9879 \\
SRResNet-6 \!+\! ShiftNet & 0.9808 & 0.9794 \\
\textbf{HighRes-net (ours)} & 0.9496 & 0.9488 \\
\textbf{HighRes-net+ (ours)} & \textbf{0.9474} & 0.9477 \\
DeepSUM & 0.9488 & \textbf{0.9474} \\
\bottomrule
\end{tabular}
\end{sc}
\end{small}
\end{center}
\vskip -0.3in
\end{table}

\subsubsection{Ablation study}

We ran an ablation study on the labeled data (1450 image sets), split in 90\% / 10\% for training and testing.
Our results suggest that more low-resolution views improve the reconstruction, plateauing after $16$ views, see Supplementary Material \ref{appendix:b}.
Another finding is that registration matters for MFSR, both in co-registering low-resolution views, and the registered loss, see Supplementary Material \ref{appendix:registration_matters}.
Finally, selecting the $k$ clearest views for fusion leads to overfitting.
A workaround is to randomly sample views with a bias for clearance, see \ref{appendix:permutation_invariance}.





\section{Discussion}
\label{sec:6.discussion}

\paragraph{On the importance of grounded detail}
Scientific and investigative application warrant a firm grounding of any prediction on real, not synthetic or hallucinated, imagery.
The PROBA-V satellite \citep{dierckx2014proba} was launched by ESA to monitor Earth's vegetation growth, water resources and agriculture.
As a form of data fusion and enrichment, multi-frame super-resolution could enhance the vision of such satellites for scientific and monitoring applications \citep{carlson1997relation, pettorelli2005using}.
More broadly, satellite imagery can help NGOs and non-profits monitor the environment and human rights \citep{cornebise2018witnessing, helber2018mapping, rudner2019multi3net, rolnick2019tackling} at scale, from space, ultimately contributing to the UN sustainable development goals. Low-resolution imagery is cheap or sometimes free, and it is frequently updated.
However, with the addition of fake or imaginary details, such enhancement would be of little value as scientific, legal, or forensic evidence.

\subsection{Future work}

Registration matters for the fusion and for the loss.
The former is not explicit in our model, and its mechanism deserves closer inspection.
Also, learning to fuse selectively with attention, would allow HighRes-net reuse all useful parts of a corrupted image.
It is hard to ensure the authenticity of detail.
It will be important to quantify the epistemic uncertainty of super-resolution for real world applications.
In the same vein, a meaningful super-resolution metric depends on the downstream prediction task.
More generally, good similarity metrics remain an open question for many computer visions tasks \citep{bruna2015super, johnson2016perceptual, isola2017image, ledig2017photo}.

\subsection{Conclusion}

We presented HighRes-net -- the first deep learning approach to MFSR that learns the typical sub-tasks of MFSR in an end-to-end fashion:
(i) \textbf{co-registration},
(ii) \textbf{fusion},
(iii) \textbf{up-sampling},
and (iv) \textbf{registration-at-the-loss}.

It recursively fuses a variable number of low-resolution views by learning a global fusion operator.
The fusion also aligns all low-resolution views with an implicit co-registration mechanism through the reference channel.
We also introduced ShiftNet-Lanczos, a network that learns to register and align the super-resolved output of HighRes-net with a high-resolution ground-truth.

Registration is vital, to align many low-resolution inputs (co-registration) and to compute similarity metrics between shifted signals.
Our experiments suggest that an end-to-end cooperative setting (HighRes-net \!+\! ShiftNet-Lanczos) improves training and test performance.
By design, our approach is fast to train and to test, with a low memory-footprint by doing the bulk of the compute (co-registration + fusion) while maintaining the low-resolution image height \& width.

There is an abundance of low-resolution yet high-revisit low-cost satellite imagery,
but they often lack the detailed information of expensive high-resolution imagery.
We believe MFSR can uplift its potential to NGOs and non-profits that contribute to the UN Sustainable Development Goals.

\makeacknowledgments





\bibliography{main}
\bibliographystyle{icml2020}





\end{document}


\twocolumn[
\icmltitle{HighRes-net: Supplementary Material}



\icmlsetsymbol{equal}{*}

\begin{icmlauthorlist}
\icmlauthor{Michel Deudon}{equal,eai}
\icmlauthor{Alfredo Kalaitzis}{equal,eai}
\icmlauthor{Israel Goytom}{mila}
\icmlauthor{Md Rifat Arefin}{mila}
\icmlauthor{Zhichao Lin}{eai}
\icmlauthor{Kris Sankaran}{mila,uom}
\icmlauthor{Vincent Michalski}{mila,uom}
\icmlauthor{Samira E. Kahou}{mila,mcg}
\icmlauthor{Julien Cornebise}{eai}
\icmlauthor{Yoshua Bengio}{mila,uom}
\end{icmlauthorlist}

\icmlaffiliation{eai}{Element AI, London, UK}
\icmlaffiliation{mila}{Mila, Montreal, Canada}
\icmlaffiliation{uom}{Universit\'{e} de Montr\'{e}al, Montreal, Canada}
\icmlaffiliation{mcg}{McGill University, Montreal, Canada}

\icmlcorrespondingauthor{Alfredo Kalaitzis}{freddie@element.ai}

\icmlkeywords{multi-frame super-resolution, super-resolution, super resolution, multi-frame, multi-image, multi image, multi frame, multi-temporal, fusion, upscaling, upsampling, up-scaling, up-sampling, remote sensing, remote-sensing, remote-sensing, dealiasing, de-aliasing, dealias, de-alias,  deep learning, registration, earth observation, satellite, satellite data, satellite imagery, computer vision, ICML, lanczos filter, lanczos, image registration, registration, homographynet, homography}

\vskip 0.3in
]



\printAffiliationsAndNotice{\icmlEqualContribution} 

\section{Experimental details}
\label{appendix:a}

We trained our models on low-resolution patches of size $64 \times 64$.
HighRes-net's architecture is described in Table \ref{tab:hrnet-archi}.
We denote by \texttt{Conv2d(in, out, k, s, p)} a \texttt{conv2D} layer with \texttt{in} and \texttt{out} input/output channels, kernels of size $k \times k$, stride $s$ and padding $p$.
We used the ADAM optimizer \citep{kingma2014adam} with default hyperparameters and trained our models on batches of size 32, for 400 epochs, using 90\% of the data for training and 10\% for validation.
Our learning rate is initialized to 0.0007, decayed by a factor of 0.97 if the validation loss plateaus for more than 2 epochs.
To regularize ShiftNet, we set $\lambda = 10^{-6}$.

\begin{table}[h]
\caption{\texttt{ResidualBlock(h)} architecture}
\label{tab:resblock-archi}
\vskip 0.10in
\begin{center}
\begin{small}
\begin{tabular}{lc}
\toprule
\textsc{layer0} & \texttt{Conv2d(in=h, out=h, k3, s1, p1)} \\
\textsc{layer1} & \texttt{PReLU} \\
\textsc{layer2} & \texttt{Conv2d(in=h, out=h, k3, s1, p1)} \\
\textsc{layer3} & \texttt{PReLU} \\
\bottomrule
\end{tabular}
\end{small}
\end{center}
\vskip -0.1in
\end{table}

\begin{table*}
\caption{\texttt{HRNet} architecture}
\label{tab:hrnet-archi}
\vskip 0.10in
\begin{center}
\begin{small}
\begin{tabular}{lcc}
\toprule
\textsc{Step} & \textsc{Layers} & \textsc{number of parameters} \\
\midrule
\textsc{encode} & \texttt{Conv2d(in=2, out=64, k3, s1, p1)} & \textsc{1216} \\
 & \texttt{PReLU} & \textsc{1} \\
 & \texttt{ResidualBlock(64)} & \textsc{73,858} \\
 & \texttt{ResidualBlock(64)} & \textsc{73,858} \\
 & \texttt{Conv2d(in=64, out=64, k3, s1, p1)} & \textsc{36,928} \\
\midrule
\textsc{fuse} & \texttt{ResidualBlock(128)} & \textsc{295,170} \\
 & \texttt{Conv2d(in=128, out=64, k3, s1, p1)} & \textsc{73,792} \\
 & \texttt{PReLU} & 1 \\
\midrule
\textsc{decode} & \texttt{ConvTranspose2d(in=64, out=64, k3, s1)} & \textsc{36,928} \\
 & \texttt{PreLU} & 1 \\
 & \texttt{Conv2d(in=64, out=1, k1, s1)} & \textsc{65} \\
\midrule
\textsc{residual} \\ \textsc{(optional)} & \texttt{Upsample(scale\_factor=3.0, mode='bicubic')} & \textsc{0} \\
\midrule
 & & \textsc{591,818 (total)} \\
\bottomrule
\end{tabular}
\end{small}
\end{center}
\vskip 0.10in
\end{table*}

Thanks to weight sharing, HighRes-net super-resolves scenes with 32 views in 5 recursive steps, while requiring less than 600K parameters.
ShiftNet has more than 34M parameters (34,187,648) but is dropped during test time.
We report GPU memory requirements in table \ref{tab:requirements} for reproducibility purposes.

\begin{table}
\caption{GPU memory requirements to train HighRes-net + ShiftNet on patches of size $64 \times 64$ with batches of size $32$, and a variable number of low-resolution frames.}
\label{tab:requirements}
\vskip 0.10in
\begin{center}
\begin{small}
\begin{sc}
\begin{tabular}{rccc}
\toprule
\# views  & 32 & 16 & 4 \\
GPU memory (GB) & 27 & 15 & 6 \\
\bottomrule
\end{tabular}
\end{sc}
\end{small}
\end{center}
\vskip -0.1in
\end{table}

\subsection{How many frames do you need?}
\label{appendix:b}

\begin{figure}
    \begin{center}
        \includegraphics[width=0.99\linewidth]{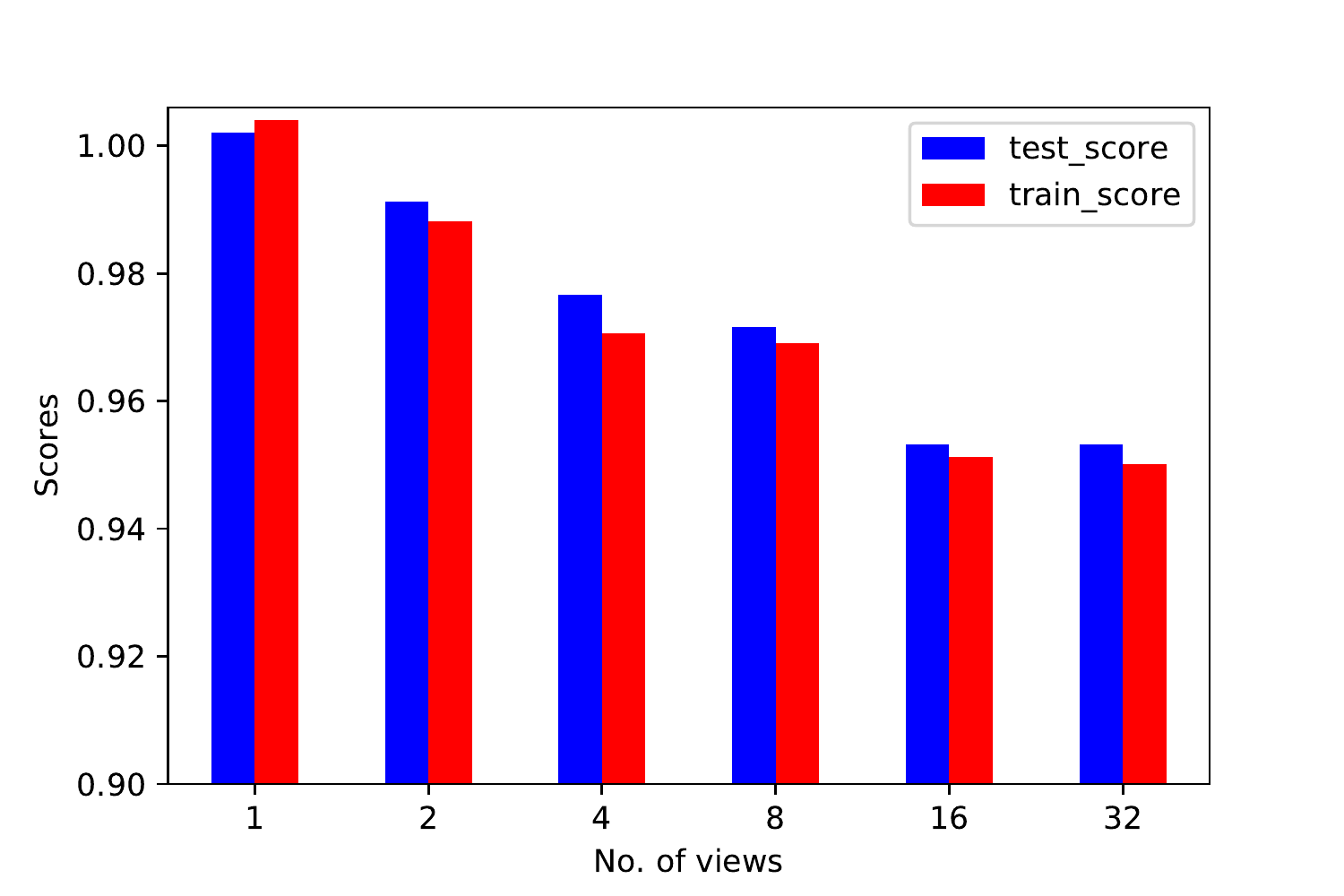}
    \end{center}
    \caption{ Public leaderboard scores vs.  nviews for HighRes-net + ShiftNet. Lower is better.
    }
    \label{fig:nviews_scores}
\end{figure}

We trained and tested HighRes-net with ShiftNet using 1 to 32 frames. With a single image, our approach performs worse than the ESA baseline. Doubling the number of frames 
significantly improves both our training and validation scores. After 16 frames, our model's performance stops increasing as show in \autoref{fig:nviews_scores}.

\subsection{Registration matters} \label{appendix:registration_matters}

\subsubsection{Registered loss}

The only explicit registration that we perform is at the loss stage, to allow the model partial credit for a solution.
This solution can be enhanced but otherwise mis-registered with respect to the ground truth.
We trained our base model HighRes-net without ShiftNet-Lanczos and observed a drop in performance as shown in Table \ref{tab:regis}.
Registration matters and aligning outputs with targets helps HighRes-net generate sharper outputs and achieve competitive results.

\begin{table}
\caption{Registration matters: Train and test scores for HighRes-net trained with and without ShiftNet-Lanczos. Lower is better.}
\label{tab:regis}
\vskip 0.10in
\begin{center}
\begin{small}
\begin{sc}
\begin{tabular}{lccc}
\toprule
& \multicolumn{2}{c}{score} \\
HighRes-net & train & test \\
\midrule
unregistered loss & 0.9616 & 0.9671 \\
registered loss & \textbf{0.9501} & \textbf{0.9532} \\
\bottomrule
\end{tabular}
\end{sc}
\end{small}
\end{center}
\vskip -0.1in
\end{table}

\subsubsection{Implicit co-registration}

The traditional practice in MFSR is to explicitly co-register the LR views prior to super-resolution \citep{tsai1984multiple, molini2019deepsum}.
The knowledge of sub-pixel miss-alignments tells an algorithm what pieces of information to fuse from each LR image for any pixel in the SR output.
Contrary to the conventional practice in MFSR, we propose implicit co-registeration by pairing LR views with a reference frame, also known as an \textit{anchor}.
In this sense, we never explicitly compute the relative shifts between any LR pair.
Instead, we simply stack each view with a chosen reference frame as an additional channel to the input.
We call this strategy \emph{implicit co-registration}.
We found this strategy to be effective in the following ablation study which addresses the impact of the choice of a reference frame aka \emph{anchor}.

We observe the median reference is the most effective in terms of train and test score. We suspect the median performs better than the mean because the median is more robust to outliers and can help denoise the LR views.
Interestingly, training and testing without a shared reference performed worse than the ESA baseline.
This shows that co-registration (implicit or explicit) matters.
This can be due to the fact that the model lacks information to align and fuse the multiple views.



\begin{table}
\caption{Scores for HighRes-net + ShiftNet-Lanczos trained and tested with different references as input. Lower is better.}
\label{tab:ref}
\vskip 0.10in
\begin{center}
\begin{small}
\begin{sc}
\begin{tabular}{lcc}
\toprule
& \multicolumn{2}{c}{score} \\
Reference & train & test \\
\midrule
no co-registration & 1.0131 & 1.0088  \\
Mean of 9 LRs & 0.9636 & 0.9690 \\
Median or 9 LRs (base) & \textbf{0.9501} & \textbf{0.9532} \\
\bottomrule
\end{tabular}
\end{sc}
\end{small}
\end{center}
\end{table}

\subsubsection{ShiftNet architecture}

ShitNet has 8 layers of \texttt{(conv2D + BatchNorm2d + ReLU)} modules.
Layer 2, 4 and 6 are followed by \texttt{MaxPool2D}.
The final output is flattened to a vector $x$ of size $32,768$.
Then, we compute a vector of size $1,024$,
\texttt{x = ReLU(fc1(dropout(x)))}.
The final shift prediction is \texttt{fc2(x)} of size $2$.
The bulk of the parameters come from \texttt{fc1}, with $32,768 \times 1,024$ weights.
These alone, account for 99\% of ShiftNet's parameters.
Adding a \texttt{MaxPool2D} on top of layer 3, 5, 7 or 8 halves the parameters of ShiftNet.

\subsection{Towards permutation invariance} \label{appendix:permutation_invariance}

A desirable property of a fusion model acting on an un-ordered set of images, is permutation-invariance:
the output of the model should be invariant to the order in which the LR views are fused.
An easy approach to encourage permutation invariant neural networks is to randomly shuffle the inputs at training time before feeding them to a model \citep{vinyals2015order}.

In addition to randomization, we still want to give more importance to clear LR views (with high clearance score), which can be done by sorting them by clearance. A good trade-off between uniform sampling and deterministic sorting by \emph{clearance}, is to sample $k$ LR views without replacement and with a bias towards higher clearance:

\begin{equation}
    p(i ~|~ C_{1},\dots,C_{k})=\frac{e^{\beta C_{i}}}{\sum_{j=1}^{k} e^{\beta C_{j}}},
\end{equation}
where $k$ is the total number of LR views, $C_{i}$ is the clearance score of LR$_{i}$ and $\beta$ regulates the bias towards higher clearance scores, 

When $\beta=0$, this sampling strategy corresponds to uniform sampling and when $\beta=+inf$, this corresponds to picking the k-clearest views in a deterministic way.
Our default model was trained with $\beta=50$ and our experiments are reported in Table \ref{tab:beta}. 

\begin{table}
\caption{scores per sampling strategy for HighRes-net + ShiftNet. Lower is better.}
\label{tab:beta}
\vskip 0.10in
\begin{center}
\begin{small}
\begin{sc}
\begin{tabular}{lccc}
\toprule
& \multicolumn{2}{c}{score} \\
Sampling strategy & train & test \\
\midrule
$\beta = \infty$  (k-clearest) & \textbf{0.9386} & 0.9687  \\
$\beta = 0$ (uniform-k) & 0.9638 & 0.9675 \\
$\beta = 50$ (base) & 0.9501 & \textbf{0.9532} \\
\bottomrule
\end{tabular}
\end{sc}
\end{small}
\end{center}
\vskip -0.10in
\end{table}

From Table \ref{tab:beta}, $\beta = \infty$ reaches best training score and worst testing score. For  $\beta = 50$ and $\beta = 0$, the train/test gap is much more reduced.
This suggests that the deterministic strategy is overfitting and randomness prevents overfitting (diversity matters).
On the other hand, $\beta = 50$  performs significantly better than $\beta = 0$ suggesting that biasing a model towards higher clearances could be beneficial i.e., clouds matter too.

\section{On the parallax effect}

The parallax $p$ is a measure of space that is inversely proportional to the distance $d$ from the object, see e.g. \citep{zeilik1998introductory}:
$$p \propto 1 / d$$

If 32 low-resolution satellite views are acquired during a single fly-over, the successive geolocations would indeed be significantly different and the parallax effect would be magnified.
This is indeed the case in aerial photography, for instance, because the imagery is captured from much closer to the ground.

With satellite imagery, one might be interested in detecting vegetation growth (PROBA-V), road networks (infrastructure), farms / ranches (agriculture), deforestation in the Amazon, or human presence and buildings.
In all these monitoring applications, the objects of interests are no more than 50m tall, e.g. trees.

The parallax effect between low-res images does not inhibit the super-resolution of the such objects. The lowest of Low Earth Orbit (LEO) altitudes for a satellite is 300 km (PROBA-V is about 800km), so the relative depth variation is at most 50m / 300,000m = 0.0033\%.
In other words, the parallax effect is imperceptible for 50m tall objects.

Here is a calculation to support this argument:
Given a point A at height 50m (distance $d_A = 300,000m - 50m$), and a point B at height 0 (distance $d_B = 300,000m$), their relative change in motion is:
$p_A / p_B = d_B / d_A = 
30 / 29.995$.
This means that if point A moves 30m, then point B moves 5 millimeters less than 30m due to parallax.
In the case of a fast LEO satellite like PROBA-V, its geolocation is accurate enough such that the translational shifts are mostly within a sub-pixel accuracy, and they almost never exceed 2 pixels.
On the ground, 2 pixels amount to a baseline length of (2 px) * (300 m/px) = 600m.
So between two images where point A (50m altitude) moved 600m, and point B (0m altitude) has moved
0.005 * 20 = 0.1 m.
Hence, the parallax effect is imperceptible for 50m tall objects. Even less so for the objects that we have underlined above, and our experimental state-of-the-art results support this claim.



\clearpage
\bibliography{main}
\bibliographystyle{icml2020}